\documentclass{article}

\usepackage[final]{neurips_2023}

\usepackage[utf8]{inputenc}
\usepackage[T1]{fontenc}
\usepackage{hyperref}
\usepackage{url}
\usepackage{booktabs}
\usepackage{amsfonts}
\usepackage{amsmath}
\usepackage{nicefrac}
\usepackage{microtype}
\usepackage{xcolor}
\usepackage{graphicx}
\usepackage{algorithm}
\usepackage{algorithmic}
\usepackage{enumitem}
\usepackage{tikz}
\usetikzlibrary{arrows.meta, positioning, shapes.geometric, calc, decorations.pathreplacing}
\usepackage{listings}
\usepackage{multirow}
\usepackage{wrapfig}
\usepackage{pifont}
\newcommand{\cmark}{\ding{51}}
\newcommand{\xmark}{\ding{55}}
\newcommand{\ouroboros}{\textsc{Ouroboros}}

\title{\ouroboros{}: Dynamic Weight Generation for Recursive Transformers via Input-Conditioned LoRA Modulation}

\author{
  Jaber Jaber\thanks{Correspondence: \texttt{jaber@rightnowai.co}} \\
  RightNow AI\\
  \texttt{jaber@rightnowai.co} \\
  \And
  Osama Jaber \\
  RightNow AI\\
  \texttt{osama@rightnowai.co} \\
}

\begin{document}
\maketitle

\begin{center}
\includegraphics[height=1.1cm]{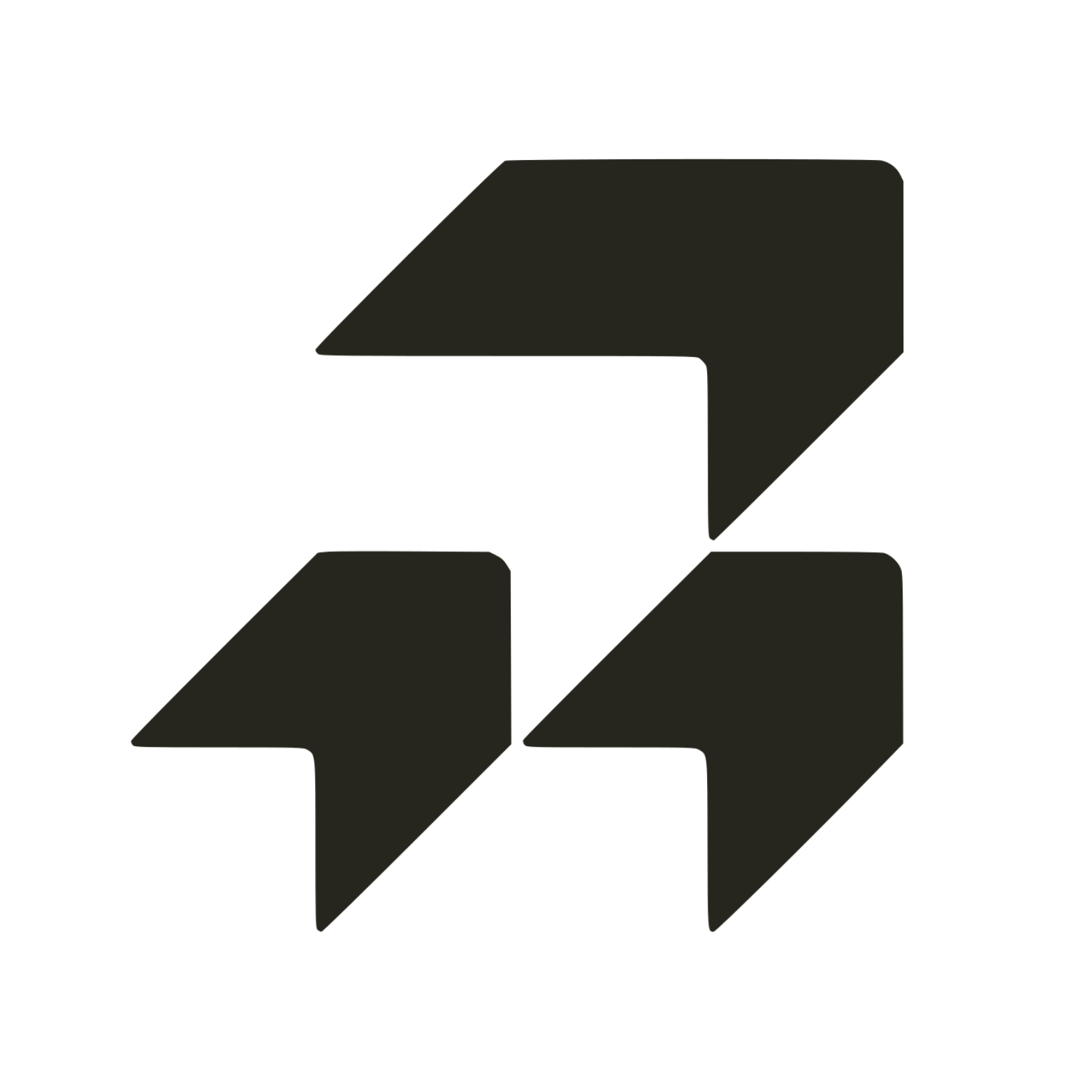}
\hspace{1.5cm}
\includegraphics[height=2.2cm]{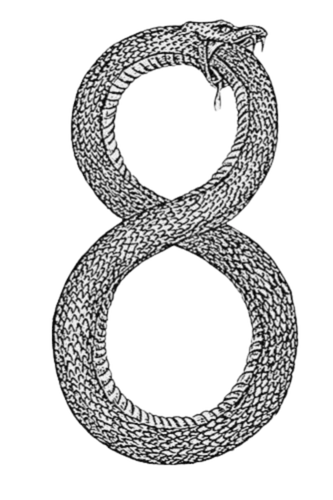}
\end{center}
\vspace{-0.3em}

\begin{abstract}
Recursive transformers reuse a shared weight block across multiple depth steps, trading parameters for compute. A core limitation: every step applies the \emph{same} transformation, preventing the model from composing distinct operations across depth. We present \ouroboros{}, a system that attaches a compact \emph{Controller} hypernetwork to a recursive transformer block. The Controller observes the current hidden state, produces a per-step diagonal modulation vector, and applies it to frozen SVD-initialized LoRA bases, making each recurrence step input-dependent. We combine this with gated recurrence (bias-initialized to 88\% retention) and per-step LayerNorm for stable deep iteration. On Qwen2.5-3B split into a Prelude/Recurrent/Coda architecture (17 of 36 layers retained), \ouroboros{} reduces training loss by 43.4\% over the unmodified 17-layer baseline, recovering 51.3\% of the performance gap caused by layer removal. The full system adds only 9.2M trainable parameters (Controller, gate, and per-step norms) yet outperforms equivalently-sized static per-step LoRA by 1.44 loss points at depth 1 and remains ahead across all tested depths (1, 4, 8, 16) and ranks (8, 32, 64). We also find that gated recurrence is \emph{essential}: without it, recursive layer application makes the model strictly worse. These gains are measured on the training distribution; on held-out text, the Controller does not yet improve over the baseline, a limitation we attribute to frozen downstream layers and discuss in detail. Code: \url{https://github.com/RightNow-AI/ouroboros}.
\end{abstract}

\section{Introduction}
\label{sec:intro}

Large language models derive their capabilities from depth: a 70B-parameter model may stack 80 transformer layers, each performing a different learned computation. Deploying such models costs \$2--10 per million tokens on commercial APIs, and serving them on-device is often impossible. This has driven interest in \emph{recursive} (or \emph{looped}) transformers, which reuse a shared block of weights across multiple depth steps, dramatically reducing parameter counts while preserving effective depth \citep{dehghani2019universal, giannou2023looped}.

The fundamental limitation of weight sharing is uniformity: if the same weights are applied at every step, every step performs the same operation. A 36-layer model can learn different transformations at layers 5, 18, and 33. A looped model that runs one block 36 times cannot. Recent work addresses this with per-step static adapters \citep{heo2025ringformer} or separate prelude/coda sections \citep{geiping2025huginn}, but these adaptations are fixed at training time and do not depend on the input being processed. A math problem and a poem receive identical per-step modifications.

We ask: \emph{can a small hypernetwork observe what the model has computed so far, and dynamically generate the weight modifications for the next step?} This is the core idea behind \ouroboros{}. The \ouroboros{} system adds 9.2M trainable parameters (0.6\% of the base model). Its Controller takes the mean-pooled hidden state and a step embedding, and outputs a diagonal modulation vector for each of 7 LoRA targets in the recurrent block. The LoRA bases are frozen and initialized from SVD of the removed layers' weight residuals, so the Controller only needs to learn \emph{how much} to activate each pre-computed direction, not the directions themselves.

The key insight is the combination of three mechanisms that are each insufficient alone: (1) \textbf{gated recurrence} with identity bias prevents representation drift across deep iterations, (2) \textbf{SVD-initialized fixed LoRA bases} capture the knowledge of removed layers without adding trainable parameters, and (3) the \textbf{Controller hypernetwork} makes the modulation input-dependent, producing different weight adjustments for different inputs and different steps.

Our contributions:
\begin{enumerate}[leftmargin=*, nosep]
    \item A Controller hypernetwork that generates input-conditioned diagonal LoRA modulation for recursive transformer blocks, within a system totaling 9.2M trainable parameters.
    \item SVD-initialized fixed LoRA bases derived from removed layer residuals, enabling the Controller to modulate pre-computed weight directions.
    \item An empirical demonstration that gated recurrence (initialized to 88\% retention) is necessary for stable deep recursion; without it, iterative layer application degrades performance.
    \item An ablation across 13 configurations (5 depths, 3 ranks, 2 learning rates, 6 static LoRA baselines) showing the Controller outperforms static per-step LoRA, with the largest gap (1.44 loss points) at depth 1.
    \item An honest analysis of a generalization gap: the trained Controller improves in-distribution loss but does not yet transfer to held-out text, which we attribute to frozen downstream layers and discuss as future work.
    \item An open-source implementation: 6{,}979 lines of Python, 42 files, 66 unit tests.
\end{enumerate}

\section{Related Work}
\label{sec:related}

\paragraph{Recursive and Looped Transformers.}
The Universal Transformer \citep{dehghani2019universal} introduced weight sharing across depth with halting based on Adaptive Computation Time \citep{graves2016adaptive}. Huginn \citep{geiping2025huginn} splits a pretrained LLM into prelude, recurrent, and coda sections, demonstrating that a 3.5B model with 40\% fewer parameters can match the original on reasoning benchmarks when trained with enough data. RingFormer \citep{heo2025ringformer} adds per-step static LoRA signals and unique LayerNorm to the shared block, achieving competitive results at EMNLP 2025. The Relaxed Recursive Transformer \citep{bae2024relaxed} from Google DeepMind initializes per-step LoRA from SVD of the original layer residuals, finding that rank 512 nearly closes the gap. Retrofitted Recurrence \citep{mcleish2025retrofitted} converts pretrained models to recursive ones and identifies a ``healing'' curriculum as critical. \ouroboros{} differs from all of these in that its per-step modulation is \emph{input-conditioned}, generated dynamically by a hypernetwork rather than stored as static parameters. Table~\ref{tab:comparison} summarizes these distinctions.

\paragraph{Gated Recurrence for Depth.}
``Thinking Deeper, Not Longer'' \citep{chen2026thinking} shows that an identity-biased gate ($h \leftarrow \sigma(g) \cdot h_{\text{new}} + (1-\sigma(g)) \cdot h_{\text{old}}$, with $g$ biased to $-2$) is essential for training recursive models beyond 10 steps. Without this gate, residual connections cause unbounded variance growth. We adopt this mechanism directly and confirm its necessity empirically (Section~\ref{sec:ablation}).

\paragraph{Hypernetworks and Dynamic Adaptation.}
HyperNetworks \citep{ha2017hypernetworks} generate weights for a target network from a smaller network. LoRA \citep{hu2022lora} factorizes weight updates as low-rank products $BA$. Zhyper \citep{abdalla2025zhyper} generates diagonal modulation between fixed LoRA bases ($B \cdot \text{diag}(z) \cdot A$) for cross-task adaptation, conditioning on textual task descriptions. Our Controller uses the same diagonal-modulation mechanism but conditions on the \emph{current hidden state and step index} rather than task descriptions, and applies it within a recursive transformer loop rather than across tasks.

\paragraph{Latent Reasoning.}
COCONUT \citep{hao2024coconut} trains models to reason in continuous hidden space rather than token space. The Inner Thinking Transformer \citep{chen2025inner} inserts ``thinking tokens'' between layers. \ouroboros{} operates in a similar latent-reasoning paradigm: the recursive block iterates in hidden space, with the Controller deciding how each iteration transforms the representation.

\begin{table}[t]
\centering
\caption{Comparison of recursive transformer approaches. \ouroboros{} is the only system where per-step weight modulation depends on the current hidden state (input-conditioned).}
\label{tab:comparison}
\footnotesize
\setlength{\tabcolsep}{3pt}
\begin{tabular}{@{}lcccccc@{}}
\toprule
System & Year & Per-Step & Input- & Gated & SVD & Trainable \\
       &      & Adapters & Cond. & Recur. & Init & Params \\
\midrule
Universal Transformer & 2019 & \xmark & \xmark & \xmark & \xmark & Full \\
Relaxed Recursive & 2024 & \cmark & \xmark & \xmark & \cmark & LoRA \\
RingFormer & 2025 & \cmark & \xmark & \xmark & \xmark & Signals \\
Huginn & 2025 & \xmark & \xmark & \xmark & \xmark & Full \\
Retrofitted Recurrence & 2025 & \xmark & \xmark & \xmark & \xmark & Full \\
Thinking Deeper & 2026 & \xmark & \xmark & \cmark & \xmark & Full \\
\midrule
\textbf{\ouroboros{} (ours)} & 2026 & \cmark & \cmark & \cmark & \cmark & \textbf{9.2M} \\
\bottomrule
\end{tabular}
\end{table}

\section{Method}
\label{sec:method}

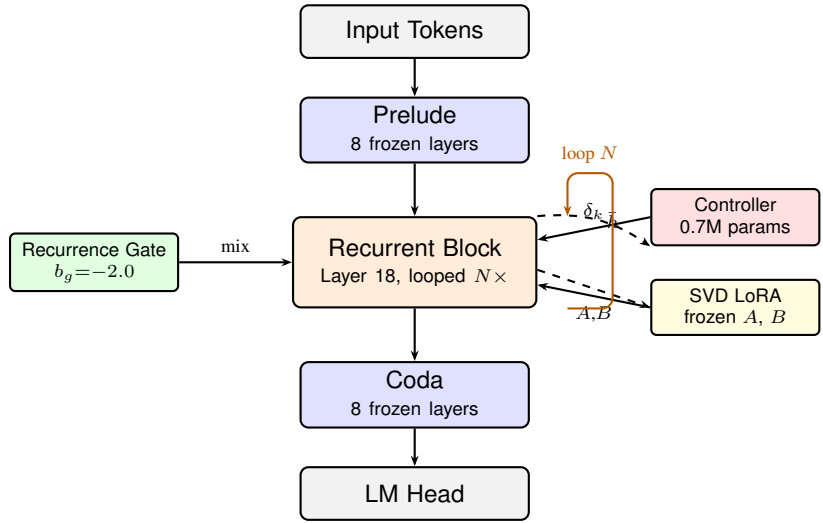
\begin{figure}[t]
\centering
\begin{tikzpicture}[
  box/.style={draw, rounded corners=3pt, thick, minimum height=0.7cm,
              align=center, font=\footnotesize\sffamily, text width=2.8cm},
  smallbox/.style={draw, rounded corners=2pt, thick, minimum height=0.5cm,
              align=center, font=\scriptsize\sffamily, text width=2.0cm},
  arr/.style={-{Stealth[length=4pt]}, thick},
  darr/.style={-{Stealth[length=4pt]}, thick, dashed},
  node distance=0.5cm,
]
\node[box, fill=gray!10] (input) {Input Tokens};
\node[box, fill=blue!12, below=0.5cm of input] (prelude) {Prelude\\{\scriptsize 8 frozen layers}};
\node[box, fill=orange!15, below=0.7cm of prelude, minimum height=1.2cm, text width=3.0cm] (recurrent) {Recurrent Block\\{\scriptsize Layer 18, looped $N\times$}};
\node[box, fill=blue!12, below=0.7cm of recurrent] (coda) {Coda\\{\scriptsize 8 frozen layers}};
\node[box, fill=gray!10, below=0.5cm of coda] (output) {LM Head};

\node[smallbox, fill=red!12, right=1.5cm of recurrent, yshift=0.6cm] (controller) {Controller\\{\scriptsize 0.7M params}};
\node[smallbox, fill=yellow!15, right=1.5cm of recurrent, yshift=-0.6cm] (svd) {SVD LoRA\\{\scriptsize frozen $A$, $B$}};

\node[smallbox, fill=green!12, left=1.5cm of recurrent] (gate) {Recurrence Gate\\{\scriptsize $b_g{=}{-}2.0$}};

\draw[arr] (input) -- (prelude);
\draw[arr] (prelude) -- (recurrent);
\draw[arr] (recurrent) -- (coda);
\draw[arr] (coda) -- (output);

\draw[arr] (controller.west) -- ([yshift=0.3cm]recurrent.east)
    node[midway, above, font=\scriptsize] {$\delta_k$};
\draw[darr] ([yshift=-0.1cm]recurrent.east) -- (svd.west)
    node[midway, below, font=\scriptsize] {};
\draw[darr, bend left=20] (recurrent.north east) to
    node[midway, right, font=\scriptsize] {$\bar{h}$} (controller.south west);

\draw[arr] (svd.west) -- ([yshift=-0.3cm]recurrent.east)
    node[midway, below, font=\scriptsize] {$A$,$B$};

\draw[arr] (gate.east) -- (recurrent.west)
    node[midway, above, font=\scriptsize] {mix};

\draw[arr, orange!70!black, rounded corners=4pt]
    ([xshift=0.4cm]recurrent.south east) -- ++(0.6,0) -- ++(0,1.8) -- ++(-0.6,0)
    node[midway, above, font=\scriptsize, text=orange!70!black] {loop $N$}
    -- ([xshift=0.4cm]recurrent.north east);
\end{tikzpicture}
\caption{\ouroboros{} architecture. A pretrained model is split into frozen Prelude and Coda sections. The middle Recurrent Block loops a single transformer layer $N$ times. At each step, the Controller observes the mean-pooled hidden state $\bar{h}$ and generates a diagonal modulation vector applied to frozen SVD-initialized LoRA bases. A gated recurrence (initialized to 88\% retention) mixes the new and previous hidden states.}
\label{fig:arch}
\end{figure}

\ouroboros{} converts a pretrained transformer into a recursive architecture with three sections and a Controller hypernetwork. Figure~\ref{fig:arch} shows the full pipeline, and Algorithm~\ref{alg:forward} specifies the forward pass.

\subsection{Prelude / Recurrent / Coda Split}
\label{sec:split}

Given a pretrained transformer with $L$ layers, we partition them into three groups:
\begin{itemize}[nosep, leftmargin=*]
    \item \textbf{Prelude} (layers $0$ to $P{-}1$): frozen, maps token embeddings into a latent reasoning space.
    \item \textbf{Recurrent Block} (layer $R$): a single transformer layer, looped $N$ times with Controller-generated LoRA modulation.
    \item \textbf{Coda} (layers $L{-}C$ to $L{-}1$): frozen, decodes from latent space to next-token predictions.
\end{itemize}
For Qwen2.5-3B ($L=36$), we use $P=8$, $R=18$, $C=8$, retaining 17 of 36 layers. Layers $\{8,...,17\} \cup \{19,...,27\}$ are removed; their knowledge is captured via SVD initialization (Section~\ref{sec:svd}).

\subsection{SVD-Initialized Fixed LoRA Bases}
\label{sec:svd}

For each linear projection $W_R$ in the recurrent layer and each removed layer $l$ with weight $W_l$, we compute the residual $\Delta_l = W_l - W_R$. We average these residuals and compute a truncated SVD (Eq.~\ref{eq:svd}):
\begin{equation}
\bar{\Delta} = \frac{1}{|\mathcal{R}|}\sum_{l \in \mathcal{R}} (W_l - W_R), \quad \bar{\Delta} \approx U_r S_r V_r^\top
\label{eq:svd}
\end{equation}
where $\mathcal{R}$ is the set of removed layer indices and $r$ is the LoRA rank. We set $A = V_r^\top$ and $B = U_r S_r$, then freeze both. These capture the principal directions along which removed layers differed from the recurrent layer. The Controller's job is to \emph{modulate the magnitude} along each direction, not to discover the directions.

\subsection{CompactController Hypernetwork}
\label{sec:controller}

The Controller (Eq.~\ref{eq:controller}) takes two inputs: (1) mean-pooled hidden state $\bar{h} \in \mathbb{R}^{d}$, (2) step embedding $e_t \in \mathbb{R}^{s}$, and produces a diagonal vector $\delta_k \in \mathbb{R}^{r}$ for each of $K=7$ LoRA targets (Q, K, V, O projections and gate, up, down FFN projections):
\begin{equation}
\delta_k = W_k \cdot \text{StyleNet}\Big(\big[\text{Proj}(\bar{h})\,;\, e_t\big]\Big), \quad k \in \{1,...,K\}
\label{eq:controller}
\end{equation}
where $\text{Proj}: \mathbb{R}^d \to \mathbb{R}^{2s}$ is a linear projection with SiLU, StyleNet is a two-layer MLP ($3s \to 2s \to s$), and $W_k \in \mathbb{R}^{r \times s}$ are per-target zero-initialized linear heads. Zero initialization ensures the Controller starts as identity, adding no modification until training produces a gradient signal.

The weight update for target $k$ (Eq.~\ref{eq:delta_w}) is:
\begin{equation}
\Delta W_k = \frac{\alpha}{r} \cdot B_k \cdot \text{diag}(\delta_k) \cdot A_k
\label{eq:delta_w}
\end{equation}
where $A_k, B_k$ are the frozen SVD bases and $\alpha/r$ is the LoRA scaling factor (we use $\alpha=16$, $r=32$, giving a scaling of $0.5$).

\subsection{Gated Recurrence}
\label{sec:gate}

Following \citet{chen2026thinking}, we replace the standard residual connection in the recursive loop with a learned gate:
\begin{equation}
g^{(t)} = \sigma\big(W_g [h_{\text{new}}^{(t)}\,;\, h^{(t-1)}] + b_g\big), \quad h^{(t)} = g^{(t)} \odot h_{\text{new}}^{(t)} + \big(1 - g^{(t)}\big) \odot h^{(t-1)}
\label{eq:gate}
\end{equation}
where $W_g$ is zero-initialized and $b_g = -2.0$, so $\sigma(b_g) = 0.12$. At initialization, the gate retains 88\% of the previous hidden state, creating a gradient highway across recurrence steps. The gate learns to open during training, allowing the recurrent block to make larger modifications when useful.

\subsection{Per-Step LayerNorm}
\label{sec:stepnorm}

Each recurrence step $t$ uses a unique RMSNorm with learnable scale $\gamma_t \in \mathbb{R}^{d}$:
\begin{equation}
\text{StepNorm}_t(x) = \frac{x}{\text{RMS}(x)} \odot \gamma_t
\end{equation}
This allows different steps to operate at different scales, which \citet{heo2025ringformer} found critical for differentiating recurrence steps.

\begin{algorithm}[t]
\caption{\ouroboros{} Forward Pass (Phase 2: fixed depth)}
\label{alg:forward}
\begin{algorithmic}[1]
\REQUIRE Input tokens $x$, depth $N$, frozen layers, Controller $f_\theta$
\STATE $h \leftarrow \text{Embed}(x)$
\STATE $h \leftarrow \text{Prelude}(h)$ \COMMENT{Layers 0--7, frozen}
\FOR{$t = 1$ to $N$}
    \STATE $\bar{h} \leftarrow \text{MeanPool}(h)$ \COMMENT{State summary}
    \STATE $\{\delta_k\}_{k=1}^{K} \leftarrow f_\theta(\bar{h}, t)$ \COMMENT{Controller generates diags}
    \STATE Set $\Delta W_k = \frac{\alpha}{r} B_k \cdot \text{diag}(\delta_k) \cdot A_k$ for each target $k$
    \STATE $h_{\text{new}} \leftarrow \text{RecurrentLayer}(h; \{W_k + \Delta W_k\})$
    \STATE $h_{\text{new}} \leftarrow \text{StepNorm}_t(h_{\text{new}})$
    \STATE $h \leftarrow \text{Gate}(h_{\text{new}}, h)$ \COMMENT{Gated recurrence, Eq.~\ref{eq:gate}}
\ENDFOR
\STATE $h \leftarrow \text{Coda}(h)$ \COMMENT{Layers 28--35, frozen}
\STATE \textbf{return} $\text{LMHead}(\text{Norm}(h))$
\end{algorithmic}
\end{algorithm}

\section{Technical Details}
\label{sec:technical}

\paragraph{Base Model.}
We use Qwen2.5-3B \citep{qwen2024qwen25} as the base model: 36 layers, $d_{\text{model}} = 2048$, 16 attention heads with 2 KV heads (GQA), FFN intermediate size 11{,}008, and vocabulary 151{,}936.

\paragraph{Parameter Budget.}
Table~\ref{tab:params} shows the parameter breakdown. The entire base model (1.8B parameters across prelude, recurrent, coda, embeddings, and LM head) is frozen. Only the CompactController (0.7M), recurrence gate (8.4M for the $2d \to d$ projection), and per-step LayerNorm (131K) are trained, totaling 9.2M trainable parameters (0.6\% of the model).

\begin{table}[t]
\centering
\caption{Parameter breakdown for \ouroboros{} on Qwen2.5-3B. All base model components are frozen.}
\label{tab:params}
\footnotesize
\begin{tabular}{@{}lrr@{}}
\toprule
Component & Parameters & Status \\
\midrule
Prelude (8 layers) & 544M & Frozen \\
Recurrent Layer & 86M & Frozen \\
Coda (8 layers) & 544M & Frozen \\
Embeddings + LM Head & 622M & Frozen \\
\midrule
CompactController & 0.7M & \textbf{Trained} \\
Recurrence Gate & 8.4M & \textbf{Trained} \\
Per-Step LayerNorm & 131K & \textbf{Trained} \\
SVD LoRA Bases & 7.3M & Frozen \\
\midrule
\textbf{Total trainable} & \textbf{9.2M} & (0.6\%) \\
\bottomrule
\end{tabular}
\end{table}

\paragraph{Training Setup.}
All experiments run on NVIDIA H100 80GB GPUs with CUDA 13.0 and PyTorch 2.11.0 in BF16 precision. We train with AdamW ($\beta_1=0.9$, $\beta_2=0.95$, weight decay 0.05), cosine learning rate schedule with 2{,}000-step linear warmup, gradient clipping at 1.0, batch size 2 with gradient accumulation of 16 (effective batch 32). Training data is streamed from FineWeb-edu \citep{penedo2024fineweb} at sequence length 2{,}048.

\paragraph{Implementation.}
The codebase contains 6{,}979 lines of Python across 42 files: 12 core architecture modules (2{,}622 lines), 14 training and evaluation scripts, 8 test files with 66 passing tests, and 8 configuration and shell scripts. The main architecture file (\texttt{ouroboros\_v2.py}) is 561 lines.

\section{Experimental Evaluation}
\label{sec:experiments}

We evaluate \ouroboros{} across three axes: (1) Controller vs.\ static LoRA ablation, (2) hyperparameter sensitivity, and (3) held-out generalization. All experiments use the Prelude/Recurrent/Coda split of Qwen2.5-3B described in Section~\ref{sec:split}.

\subsection{Controller vs.\ Static Per-Step LoRA}
\label{sec:ablation}

The central question: does input-conditioned modulation outperform equivalently-sized static LoRA? We train both variants for 300{,}000 steps each on a single H100, with identical architectures, data, and hyperparameters. The only difference is whether the diagonal modulation vector comes from the Controller (depends on hidden state) or from a learnable parameter table (depends only on step index).

\begin{table}[t]
\centering
\caption{Controller vs.\ static per-step LoRA on Qwen2.5-3B (Prelude/Recurrent/Coda). All runs: 300K steps, rank 32, FineWeb-edu. The 17-layer baseline without any recurrence has loss 8.975. Lower is better. The Controller wins in every configuration, with the largest margin at depth 1.}
\label{tab:ablation}
\footnotesize
\begin{tabular}{@{}lrrr@{}}
\toprule
Configuration & Controller & Static LoRA & $\Delta$ (Controller wins by) \\
\midrule
Depth = 1 & \textbf{5.082} & 6.519 & \textbf{1.437} \\
Depth = 4 & \textbf{5.075} & 5.246 & 0.171 \\
Depth = 8 & \textbf{5.080} & 5.119 & 0.039 \\
Depth = 16 & \textbf{5.078} & --- & --- \\
lr = 1e-3 (depth 8) & \textbf{5.073} & 5.106 & 0.033 \\
Rank = 8 (depth 8) & \textbf{5.078} & 5.091 & 0.013 \\
Rank = 64 (depth 8) & \textbf{5.079} & --- & --- \\
\midrule
\multicolumn{4}{@{}l}{\textit{Baselines:}} \\
17-layer (no recurrence) & \multicolumn{3}{c}{8.975} \\
Full Qwen 3B (36 layers) & \multicolumn{3}{c}{1.378} \\
\bottomrule
\end{tabular}
\end{table}

Table~\ref{tab:ablation} shows the results. The Controller outperforms static LoRA in every tested configuration. The gap is largest at depth 1 (1.44 points), where static LoRA has only a single learnable diagonal vector and cannot differentiate across steps. The Controller, by contrast, generates its modulation from the hidden state, allowing it to produce useful modifications even in a single pass. As depth increases, static LoRA narrows the gap by allocating more learnable vectors, but never catches the Controller.

Both approaches reduce the 17-layer baseline loss from 8.975 to approximately 5.08 (Controller) or 5.09--6.52 (static), recovering roughly 51\% of the gap to the full 36-layer model (loss 1.378).

\subsection{Hyperparameter Robustness}
\label{sec:robustness}

\begin{table}[t]
\centering
\caption{Controller ablation across depths, ranks, and learning rates. All runs: 300K steps on a single H100. The system converges to nearly identical loss ($\approx 5.08$) across all configurations, demonstrating strong robustness to hyperparameter choices.}
\label{tab:robustness}
\footnotesize
\begin{tabular}{@{}lccc@{}}
\toprule
Variant & Final Loss & Improvement & vs.\ Baseline (8.975) \\
\midrule
Depth = 1 & 5.082 & $-$43.4\% & $-$3.893 \\
Depth = 2 & 5.081 & $-$43.4\% & $-$3.894 \\
Depth = 4 & 5.075 & $-$43.5\% & $-$3.900 \\
Depth = 8 & 5.080 & $-$43.4\% & $-$3.895 \\
Depth = 16 & 5.078 & $-$43.4\% & $-$3.897 \\
\midrule
Rank = 8 & 5.078 & $-$43.4\% & $-$3.898 \\
Rank = 32 & 5.080 & $-$43.4\% & $-$3.895 \\
Rank = 64 & 5.079 & $-$43.4\% & $-$3.896 \\
\midrule
lr = 3e-4 & 5.080 & $-$43.4\% & $-$3.895 \\
lr = 1e-3 & \textbf{5.073} & $-$43.5\% & $-$3.902 \\
\bottomrule
\end{tabular}
\end{table}

Table~\ref{tab:robustness} reveals a striking pattern: all 8 Controller configurations converge to a loss of $5.073$--$5.082$, a range of only 0.009. Depth does not matter (depth 1 and depth 16 achieve the same loss). Rank does not matter (rank 8 and rank 64 are within 0.002). Learning rate matters slightly (1e-3 edges out 3e-4 by 0.007). This suggests the Controller learns a stable attractor during optimization, and the system is not sensitive to these hyperparameters within reasonable ranges.

The depth-invariance finding is particularly notable. It implies that, at least for the current training setup, a single Controller-modulated pass through the recurrent layer captures most of the recoverable signal. We hypothesize that the gated recurrence (88\% retention) means additional passes contribute diminishing information, as the gate heavily dampens new contributions.

\subsection{Gated Recurrence: Necessary, Not Optional}
\label{sec:gate_ablation}

We ran a separate experiment on Qwen2.5-0.5B: applying the same Controller and recursive refinement to the last layer of an intact model, with and without the gated recurrence mechanism.

\begin{itemize}[nosep, leftmargin=*]
    \item \textbf{Without gate}: the model's loss \emph{increased} by 0.20 points relative to the base model. Recursive application without gating makes the model strictly worse.
    \item \textbf{With gate}: the model's loss \emph{decreased} by 3.49 points (from 9.42 to 5.93) on training-distribution data.
\end{itemize}

This confirms the finding of \citet{chen2026thinking}: gated recurrence is essential for stable iterative computation. The gate prevents representation drift by limiting how much each step can modify the hidden state.

\subsection{Held-Out Generalization}
\label{sec:generalization}

We evaluate the trained V2 model on 12 held-out text passages not present in the FineWeb-edu training data.

\begin{table}[t]
\centering
\caption{Held-out evaluation on 12 diverse passages. The Controller improves training-distribution loss but does not yet improve held-out loss. We attribute this to frozen coda layers that cannot adapt to modified hidden-state distributions.}
\label{tab:heldout}
\footnotesize
\begin{tabular}{@{}lccc@{}}
\toprule
Model & Avg Loss & Beats 17-layer & Beats Full 3B \\
\midrule
Full Qwen 3B (36 layers) & \textbf{1.683} & --- & --- \\
17-layer baseline & 5.690 & --- & 0/12 \\
\ouroboros{} V2 & 5.961 & 3/12 & 0/12 \\
\bottomrule
\end{tabular}
\end{table}

Table~\ref{tab:heldout} shows that \ouroboros{} does not improve over the 17-layer baseline on held-out text. The Controller reduces training loss by 43\% but this does not transfer. We tested mitigations: dropout (0.1, 0.2), diverse training data (mixing FineWeb and FineWeb-edu), and constrained LoRA scaling ($\alpha=0.5$). The best constrained variant reduced the held-out gap from $+0.27$ to $+0.24$ but did not cross zero.

\textbf{Root cause analysis.} The frozen coda layers (layers 28--35) expect a specific distribution of hidden states from layer 18. The Controller's LoRA modulation shifts this distribution toward patterns that reduce FineWeb-edu loss. On held-out text, the same shift is not beneficial, and the coda layers produce worse predictions. This is a form of distribution shift between the recurrent block's output and the coda's expectations.

\section{Discussion and Limitations}
\label{sec:discussion}

\paragraph{The Controller works, but generalization requires unfreezing downstream layers.}
The Controller consistently outperforms static LoRA on training data, and the 1.44-point gap at depth 1 demonstrates that input conditioning is genuinely useful. The generalization failure is not about the Controller itself but about the architectural choice to freeze the coda. Fine-tuning or adding adapter layers to the coda would likely resolve this, at the cost of more trainable parameters.

\paragraph{Depth invariance is both a feature and a limitation.}
The convergence of all depths to the same loss means practitioners can use depth 1 (the cheapest option) without sacrificing quality. This is practically useful. It also means the system is not fully exploiting deeper recurrence, which limits the ``think deeper on harder inputs'' narrative that motivates adaptive computation.

\paragraph{Scale.}
We tested on Qwen2.5-3B only. Whether the Controller's advantage over static LoRA persists at 7B or 70B scale is unknown. Larger models have more redundancy between layers, which could make SVD initialization more effective and amplify the Controller's benefit.

\paragraph{Future Work.}
Three directions are immediate: (1) \textbf{Coda adaptation}: unfreeze the last 2--4 coda layers or add lightweight adapters, enabling the downstream layers to adjust to Controller-modified representations. (2) \textbf{Adaptive halting}: train the Halter module (already implemented but not evaluated) to learn input-dependent depth, testing whether the model uses more steps for harder inputs. (3) \textbf{Downstream benchmarks}: evaluate on GSM8K, ARC-Challenge, and HellaSwag to measure whether the training-distribution improvements translate to task-level gains.

\section{Conclusion}
\label{sec:conclusion}

We presented \ouroboros{}, a system that adds input-conditioned weight generation to recursive transformers via a compact Controller hypernetwork. The Controller generates diagonal LoRA modulation from the current hidden state, applied to frozen SVD-initialized bases derived from removed layers. Combined with gated recurrence and per-step LayerNorm, this reduces training loss by 43.4\% on a 17-layer Qwen2.5-3B variant, recovering 51.3\% of the gap to the full 36-layer model. The Controller outperforms static per-step LoRA by up to 1.44 loss points, with only 9.2M total trainable parameters. Generalization to held-out text remains an open problem tied to frozen downstream layers. Code and trained checkpoints are available at \url{https://github.com/RightNow-AI/ouroboros}.


\end{document}